\pdfoutput=1
\documentclass[11pt]{article}

\usepackage{EMNLP2022}

\usepackage{times}
\usepackage{latexsym}

\usepackage[T1]{fontenc}
\usepackage[utf8]{inputenc}

\usepackage{microtype}

\usepackage{microtype}
\usepackage{amsmath, amssymb}
\usepackage{booktabs}
\usepackage{graphicx}
\usepackage{multirow}
\usepackage{float, subfig}

\usepackage{bm}
\usepackage{algorithm}
\usepackage{pifont}
\usepackage{xspace}

\newcommand{\methodname}{ACA}

\title{Learning Robust Representations for Continual  Relation \\ Extraction via Adversarial Class Augmentation}

\author{
Peiyi Wang$^1$$\footnotemark[1]$ \quad Yifan Song$^1$$\footnotemark[1]$ \quad Tianyu Liu$^2$ \quad Binghuai Lin$^2$ \\ \textbf{Yunbo Cao$^2$ \quad Sujian Li$^1$ \quad Zhifang Sui$^1$}
\\ 
$^1$ MOE Key Laboratory of Computational Linguistics, Peking University, China  \\
$^2$ Tencent Cloud Xiaowei \\
 {wangpeiyi9979@gmail.com}; {\{yfsong, lisujian, szf\}@pku.edu.cn}\\
 {\{rogertyliu, binghuailin, yunbocao\}@tencent.com;} 
}

\begin{document}
\maketitle

\renewcommand{\thefootnote}{\fnsymbol{footnote}}
\begin{abstract}
Continual relation extraction (CRE) aims to continually learn new relations from a class-incremental data stream. CRE model usually suffers from catastrophic forgetting problem, i.e., the performance of old relations seriously degrades when the model learns new relations.
Most previous work attributes catastrophic forgetting to the corruption of the learned representations as new relations come, with an implicit assumption that the CRE models have adequately learned the old relations.
In this paper, through empirical studies we argue that this assumption may not hold, and an important reason for catastrophic forgetting is that the learned representations do not have good robustness against the appearance of analogous relations in the subsequent learning process.
To address this issue, we encourage the model to learn more precise and robust representations through a simple yet effective adversarial class augmentation mechanism (ACA), which is easy to implement and model-agnostic.
Experimental results show that ACA can consistently improve the performance of state-of-the-art CRE models on two popular benchmarks. Our code is available at \url{https://github.com/Wangpeiyi9979/ACA}.

\end{abstract}
\footnotetext[1]{Equal contribution.}
\renewcommand{\thefootnote}{\arabic{footnote}}

\section{Introduction}
\begin{figure}[t]
    \centering
    \includegraphics[width=\linewidth]{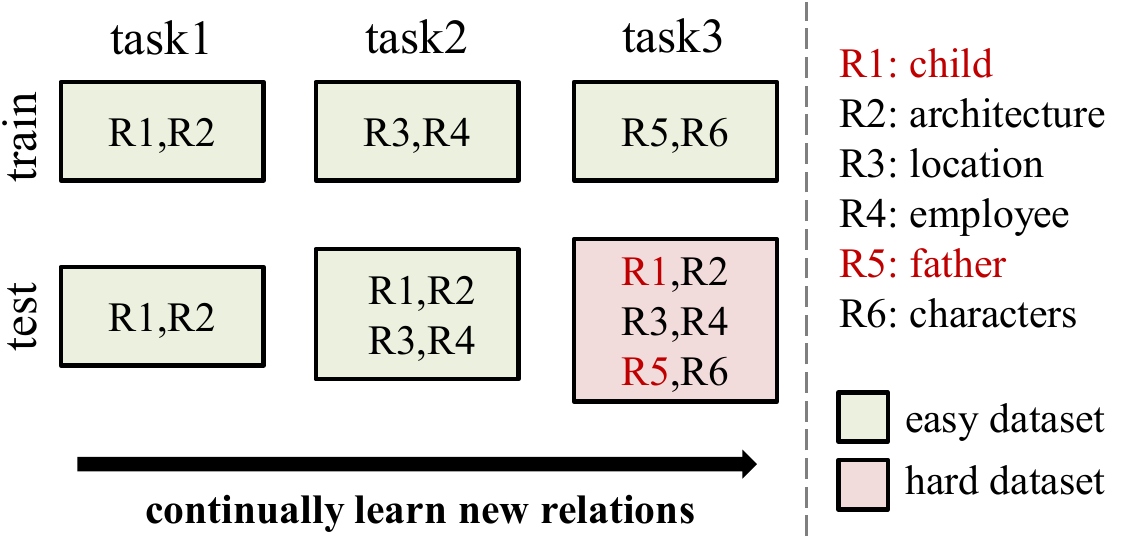}
    \caption{A demonstration for continual relation extraction with three tasks where each task involves two new relations. The learned representations from the easy training task can not handle the hard testing data, which contains analogous relations inherently hard to distinguish, e.g., ``child'' and ``father''.}
    \label{fig:intro}
\end{figure}

Relation extraction (RE) aims to detect the relation of two given entities in a sentence.
Traditional RE models are trained on a fixed dataset with a predefined relation set, which cannot handle the real-life situation where new relations are constantly emerging.
To this end, continual relation extraction (CRE) \cite{wang2019eamar, han2020emar, cui2021rpcre, zhao2022crl, wang2022less} is introduced.
As shown in Figure \ref{fig:intro}, CRE is formulated as a class-incremental problem, which trains the model on a sequence of tasks.
In each task, the model needs to learn some new relations and is evaluated on all seen relations.
Like other continual learning systems, CRE models also suffer from catastrophic forgetting, i.e., the performance of previously learned relations seriously degrades when learning new relations.

The mainstream research in CRE \cite{han2020emar,cui2021rpcre,zhao2022crl, wang2022less} mainly attributes catastrophic forgetting to the corruption of the learned knowledge as new tasks come.
To this end, a variety of sophisticated rehearsal-based mechanisms are introduced to better retain or recover the knowledge, such as relation prototypes \cite{han2020emar, cui2021rpcre}, curriculum-meta learning \cite{wu2021curriculum}, contrastive replay and knowledge distillation \cite{zhao2022crl}.
All these methods implicitly assume that the model has adequately learned old relations.
However, in this paper, we find that this assumption may not hold.

With a series of empirical studies, we observe that catastrophic forgetting mostly happens on some specific relations, and significant performance degradation tends to occur when their analogous relations appear.
Based on our observations, we find another reason for catastrophic forgetting, i.e., \emph{CRE models do not learn sufficiently robust representations of relations in the first place due to the relatively easy training task}.
Taking ``child'' in Figure \ref{fig:intro} as an example, 
% when first learning the ``child'' relation in task1, 
because of the absence of hard negative classes in task 1, the CRE model tends to rely on spurious shortcuts, such as entity types, to identify ``child''.
Although the learned imprecise representations can handle the testing set of task 1 and task 2, 
they are not robust enough to distinguish ``child'' from its analogous relation (``father'') in task 3.
Therefore, the performance of ``child'' will decrease significantly when ``father'' appears.
In contrast, relations such as ``architecture'' still perform well in task 3 because their analogous relations have not yet appeared.

Recently, adversarial data augmentation emerges as a strong baseline to prevent models from learning spurious shortcuts from the easy dataset \citep{Volpi_2018_adv, zhao_2020_maximumentropy, hendrycks_2020_augmix}.
Inspired by these work, we introduce a simple yet effective Adversarial Class Augmentation (ACA) mechanism to improve the robustness of the CRE model.
Concretely, ACA utilizes two class augmentation methods, namely hybrid-class augmentation and reversed-class augmentation, to build hard negative classes for new tasks.
When a task arrives, ACA jointly trains new relations with adversarial augmented classes to learn robust representations.
Note that our method is orthogonal to all previous work:
ACA focuses on learning knowledge of newly emerging relations better, while previous methods are proposed to retain or recover learned knowledge of old relations\footnote{The proposed method can be viewed as a ``precaution'' that takes place in the current task to mitigate the catastrophic forgetting on the analogous relations in the subsequent tasks. While the prior work is more like a ``remedy'' for the current task to recall the already learned knowledge in the past tasks.}.
Therefore, incorporating ACA into previous CRE models can further improve their performance.

We summarize our contributions as:
\textbf{1)} we conduct a series of empirical studies on two strong CRE methods and observe that catastrophic forgetting is strongly related with the existence of analogous relations.
\textbf{2)} we find an important reason for catastrophic forgetting in CRE which is overlooked in all previous work: the CRE models suffer from learning shortcuts to identify new relations, which are not robust enough against the appearance of their analogous relations.
\textbf{3)} we propose an adversarial class augmentation mechanism to help CRE models learn more robust representations. Experimental results on two benchmarks show that our method can consistently improve the performance of two state-of-the-art methods.
% \begin{itemize}
%     \item we conduct a series of empirical study to analyze catastrophic forgetting in CRE, and observe that catastrophic forgetting nearly only happens on some specific relations, and easily happens when their analogous relations appears in subsequent tasks.
%     \item we find an important reason for catastrophic forgetting in CRE that all previous works overlooked: the CRE models suffer from learning shortcuts to identify new relations, which are not robust enough against the appearance of their analogous relations.
%     \item we propose an adversarial class augmentation mechanism to help CRE models learn more robust representations. Experimental results on two benchmarks show that our method can consistently improve the performance of two state-of-the-art models.
% \end{itemize}

\section{Related Work}
\paragraph{Relation Extraction}

Conventional Relation Extraction (RE) focuses on extracting the predefined relation of two given entities in a sentence.
Recently, a variety of deep neural networks (DNN) have been proposed for RE, mainly including:
\textbf{1)} Convolutional or Recurrent neural network (CNN or RNN) based methods \cite{dossantos_2015_classifying,wang_2016_relationa,xiao_2016_semantic, liu2019reet}, which can effectively extract textual features.
\textbf{2)} Graph neural network (GNN) based methods \cite{xu_2015_semantica,xu_2016_improveda,cai_2016_bidirectionala,mandya_2020_contextualised}, which jointly encode the sentence with lexical features.
% , e.g., dependency path.
\textbf{3)} Pre-trained language model (PLM) based methods \cite{baldinisoares_2019_matching,peng_2020_learning}, which achieve state-of-the-arts on RE task.

\paragraph{Continual Learning}
Continual Learning (CL) aims to continually accumulate knowledge from a sequence of tasks \cite{delange_2019_continual}.
A major challenge of CL is catastrophic forgetting, i.e., the performance of previously learned tasks seriously drops when learning new tasks. 
To this end, prior CL methods can be roughly divided into three groups:
\textbf{1)} Rehearsal-based methods \cite{rebuffi_2017_icarl,wu_2019_large} maintain a memory to save instances of previous tasks and replay them during training of new tasks.
\textbf{2)} Regularization-based methods \cite{kirkpatrick_2017_overcoming,aljundi_2018_memory} add constraints on the weights important to old tasks.
\textbf{3)} Architecture-based methods \cite{mallya_2018_packnet,qin_2021_bns} dynamically change model architectures to learn new tasks and prevent forgetting old tasks.
Recently, rehearsal-based methods have been proven to be the most effective for NLP tasks, including relation extraction.
We focus on the rehearsal-based methods for CRE in this paper.

\paragraph{Shortcuts Learning Phenomenon}
Shortcuts learning phenomenon denotes that DNN models tend to learn unreliable shortcuts in datasets, leading to poor generalization ability in real-world applications \cite{lai-etal-2021-machine}.
Recently, researchers have revealed the shortcut learning phenomenon for different kinds of language tasks,
such as natural language inference \cite{he-etal-2019-unlearn}, information extraction \cite{wang2021behind}, reading comprehension \cite{lai-etal-2021-machine} and question answering \cite{mudrakarta-etal-2018-model}.
\citet{geirhos2020shortcut} point out that shortcuts learning phenomenon happens because the ``Principle of Least Effort'' \cite{kingsley1972human}, i.e., people (also animal and machine) will naturally minimize the amount of effort to solve tasks.
Recently, data augmentation \cite{tu-etal-2020-empirical} and adversarial training \cite{stacey-etal-2020-avoiding} are used to alleviate shortcuts learning phenomenon with synthesized data.
To the best of our knowledge, we are the first work to analyze the catastrophic forgetting in CRE from the perspective of shortcuts learning, and propose an adversarial data augmentation method to alleviate it.

\section{Task Formulation}

In CRE, the model is trained on a sequence of tasks $(T_1, T_2, ..., T_k)$.
Each task $T_i$ can be represented as a triplet $(R_i, D_i, Q_i)$, where $R_i$ is the set of new relations, $D_i$ and $Q_i$ are the training and  testing set, respectively.
Every instance $(x_j, y_j) \in D_i \cup Q_i$ belongs to a specific relation $y_j \in R_i$.
The goal of CRE is to continually train the model on new tasks to learn new relations, while avoiding forgetting of previously learned ones.
More formally, in the $i$-th task, the model learns new relations $R_i$ from $D_i$, and should be able to identify all seen relations, i.e., the model will be evaluated on the all seen testing sets
$\bigcup_{j=1}^iQ_j$.
To alleviate catastrophic forgetting in CRE, previous work \cite{cui2021rpcre,han2020emar, zhao2022crl, wang2022less} adopts a memory to store a few typical instances (e.g., $10$) for each old relation.
In the subsequent training process, instances in the memory will be replayed to alleviate the catastrophic forgetting.

\section{Catastrophic Forgetting in CRE: Characteristics and Cause}
\label{sec:pilot}

\begin{table}[t]
\centering
\small
\scalebox{0.95}{
    \begin{tabular}{lccccl}
    \toprule
    \textbf{Model} & \textbf{ID}   & \textbf{FR (\%)} & \textbf{MS} & \textbf{F1} & \textbf{F1$^*$} ($\Delta$) \\
    \midrule
    \multirow{3}{*}{\rotatebox[origin=c]{45}{EMAR}} 
    % & G1 & 0.9 & 0.41 & 97.2 & 98.6 ({$\uparrow$1.4}) \\
    % &  G2  &  2.2 & 0.45 & 91.8 & 95.1 ({$\uparrow$3.3})  \\
    % &  G3  &  4.6 & 0.54 & 84.2 & 90.4 ({$\uparrow$6.2})  \\
    % & G4  &  6.7 & 0.57 & 77.5 & 86.7 ({$\uparrow$9.2}) \\
    % & G5  &  11.0 & 0.64 & 65.6 & 78.7 ({$\uparrow$13.1}) \\
   & G1  & 1.3 & 0.42 & 95.4 & 97.4 ({$\uparrow$ 2.0}) \\
    & G2  &  4.5 & 0.53 & 84.6 & 90.9 ({$\uparrow$ 6.3})  \\
    & G3  &  9.4 & 0.62 & 69.8 & 81.5 ({$\uparrow$ 11.7})  \\
    \midrule
   \multirow{3}{*}{\rotatebox[origin=c]{45}{RP-CRE}} 
%   & G1  & 0.8 & 0.42 & 97.4 & 98.7 ({$\uparrow$1.3}) \\
%     & G2  &  2.2 & 0.44 & 91.6 & 95.1 ({$\uparrow$3.5})  \\
%     & G3  &  4.8 & 0.55 & 83.8 & 91.0 ({$\uparrow$7.2})  \\
%     & G4  &  7.1 & 0.57 & 76.9 & 86.1 ({$\uparrow$9.2}) \\
%     & G5  &  11.5 & 0.64 & 64.9 & 78.7 ({$\uparrow$13.8}) \\
   & G1  & 1.2 & 0.42 & 95.5 & 97.4 ({$\uparrow$ 1.9}) \\
    & G2  &  4.7 & 0.53 & 83.8 & 90.8 ({$\uparrow$ 7.0})  \\
    & G3  &  9.9 & 0.63 & 69.5 & 81.5 ({$\uparrow$ 12.0})  \\
    \bottomrule
    \end{tabular}
}
\caption{ We divide relations of FewRel into three groups according to their forgetting rate (FR). ``MS'' is short for \textit{max similarity}. F1 and F1$^*$ are the Macro-F1 scores of EMAR/RP-CRE and the supervised model which trains all data together, respectively. $\Delta$ is the performance gap between two CRE models and the supervised model.}
\label{tab:pilot_error}

\end{table}

In this section, we conduct a series of empirical studies on two state-of-the-art CRE models, namely EMAR \cite{han2020emar} and RP-CRE \cite{cui2021rpcre}, and two benchmarks, namely FewRel and TACRED, to analyze catastrophic forgetting in CRE.
Please refer to Section \ref{exp_setup} for details of these two benchmarks and two CRE models.

\subsection{Characteristics of Catastrophic Forgetting}
% We first compute the forgetting rate of each relation on $2$ benchmarks.
% Figure \ref{x} shows that the forgetting rate varies a lot across different relations on both benchmarks.
% To explore why there exists huge forgetting rate gap among different relations, 
We use \textit{Forgetting Rate} (FR) \cite{chaudhry_2018_riemannian, chaudhry_2018_efficient} to measure the average forgetting of a relation.
% , similar with previous works in computer vision . 
Assuming that relation $r$ appears in task $i$, the FR of $r$ after the model has finished the tasks sequence $(T_1,  ... ,T_i, ..., T_k)$ is defined as:
% been trained on the task $k\ (k>i)$ is defined as:
\begin{gather}
    FR_{r}=\frac{1}{k-i}\sum^{k}_{j=i+1} pd_{r}^{j} \\
    pd_{r}^{j}=\mathop{\rm{max}}_{l\in\{i,...,j-1\}} F1_{r}^{l}-F1_{r}^{j},
\end{gather}
% where $f_{r}^{j}$ is the forgetting of $r$ after the model learning on task $j$ and $a_{r}^{t}$ is the F1 score of $r$ in task $t$.
where $pd_{r}^{j}$ and $F1_{r}^{j}$ are the performance degradation and F1 score of $r$ after the model trains on task $j$, respectively.
The sequence length $k$ is $10$ for both FewRel and TACRED.

We  divide all relations into three equal-sized groups based on their FR from small to large.
As shown in Table \ref{tab:pilot_error}, relations in G1 hardly suffer from forgetting as FR is just $1.3\%$ and $1.2\%$ on EMAR and RP-CRE, respectively.
In contrast, relations in G3 have catastrophic forgetting and the FR is close to $10\%$ for both models.
A similar tendency is observed on TACRED (please refer to Appendix \ref{app:pilot_tacred} for details).
To explore why FR varies widely among different relations, 
we dive into the results of two CRE models and ask two questions.

% To better describe these phenomenons, we first introduce $3$ concepts:
% (a) \textbf{\textit{pair similarity}}: for two relations, the pair similarity is the cosine similarity of their definition representation.
% (b) \textbf{\textit{relation similarity}}: for a relation, relation similarity is the max \textit{pair similarity} between it and all other relation in the dataset.
% (c) \textbf{\textit{analogous relations}}: for a relation, its analogous relations are the relations with top $5$ \textit{pair similarity} in the dataset.

% \paragraph{Observation 1: The degree of forgetting is related to the similarity of relations}
\paragraph{Where catastrophic forgetting happens?}
With careful comparison between G1 and G3, we find that relations in G3 seem to have analogous relations in the dataset.
For example, ``mother'' belongs to G3, and there are its semantically analogous relations, such as ``spouse'', in the dataset.
To confirm our finding,
we first define the similarity for a pair of relations as the cosine distance of their prototypes, i.e., the mean vanilla BERT sentence embedding of all corresponding instances.
Then, for a certain relation, we compute its \textit{max similarity} (MS) to all the other relations in the dataset.
As shown in Table \ref{tab:pilot_error},
MS of G3 is significantly greater than that of G1, indicating that the catastrophic forgetting mostly happens on relations with large MS\footnote{Appendix \ref{rdg} provides more details of relations in G1/G3.}.
Besides, as shown in the last two columns (F1 and F1$^* (\Delta)$) of Table \ref{tab:pilot_error}, we also observe that the performance gap between CRE models and the supervised model significantly grows as MS increases, showing that CRE poses a more serious challenge to identify the relations with large MS.

\begin{table}[t]
\centering
\small
\scalebox{1}{
    \begin{tabular}{lcccc}
    \toprule
    \multirow{2}{*}{\textbf{Models}} & \multicolumn{2}{c}{\textbf{FewRel}} & \multicolumn{2}{c}{\textbf{TACRED}} \\
    \cmidrule(r){2-3}  \cmidrule(r){4-5}
      & \#CF & \#SIM & \#CF & \#SIM \\
    \midrule
    EMAR & 666 & 614 (92\%) & 766 & 690 (90\%)\\
    RP-CRE & 781 & 688 (88\%) & 949 & 773 (81\%)\\
    \bottomrule
    \end{tabular}
}
\caption{Analysis of catastrophic forgetting on $50$ different runs. ``\#CF'' denotes the total number of catastrophic forgetting cases, and ``\#SIM'' are bad cases accompanied by the appearance of top-$5$ analogous relations.}
\label{tab:aofb}
\end{table}

\paragraph{When catastrophic forgetting happens?}
We also observe that the performance of the relations with high FR always has a sudden drop in some tasks. 
To explore the characteristic of the task with severe performance drop, we run two CRE models on $50$ different task sequences, and record all the bad cases where catastrophic forgetting happens (the F1 scores of a relation degrades greater than $10$ points after the model learns a new task).
Given a certain relation $r$ and its corresponding bad cases, we mark cases where exist top-$5$ most similar relations of $r$.
As shown in Table \ref{tab:aofb}, we observe that more than $80\%$ bad cases on both benchmarks are related to the appearance of top-$5$ most similar relations.
Taking relation ``mother'' in FewRel dataset as an example, more than $90\%$ bad cases contain relation ``spouse'', which has the top-$1$ similarity with ``mother''\footnote{Details of some bad cases can be founded in Appendix \ref{app:sud_drop}.}.
These results show that {{for a relation that suffers catastrophic forgetting, significant performance degradation is usually accompanied by the appearance of their analogous relations.}}.

% \paragraph{Observation 2: The performance of relations with higher \textit{max similarity} lags further behind the upper bound}
% As shown in the last column of Table \ref{tab:pilot_error}, we also compute the Macro-F1 of each group under conventional supervised RE setting, which is generally seen as the upper bound of CRE.
% % In the column 3 and 6 of Table \ref{tab:pilot_error}, 
% We notice that as MS increases, the average performance of the upper bound method also decreases.
% This phenomenon may not be surprising because relations with larger MS are naturally difficult to distinguish.
% However, we further observe that the performance gap between EMAR/RP-CRE and the upper bound significantly grows as MS increases.
% This observation indicates that the effect of similarity between relations is magnified in the setting of continual learning.

\subsection{Cause of Catastrophic Forgetting}
\label{sec:reason}
All of the previous CRE works attribute the catastrophic forgetting to the corruption of the learned knowledge during the continual learning process, with the assumption that the CRE models have adequately learned the previous relations.
However, we argue that this assumption may not hold.
In CRE, models are continually trained on a sequence of stand-alone easy training datasets, where each dataset usually only consists of very few new relations without analogous relations appearing together.
In contrast, CRE models are evaluated on the much harder testing dataset of all seen relations, which usually contains analogous relations.
\citet{he-etal-2019-unlearn,karimi-mahabadi-etal-2020-end,10.5555/3524938.3525581} find that the deep neural network tends to learn spurious shortcuts in the simple training dataset to make the decision, leading to poor generalization.
Therefore, we point out another important reason for the performance degradation of learned relations:
\emph{the CRE models suffer from learning shortcuts in the easy training dataset to identify relations, which are not robust enough against the appearance of their analogous relations.}
This reason can well explain our observed phenomena, i.e, catastrophic forgetting mostly happens on some specific relations with analogous relations in the subsequent tasks, and significant performance degradation nearly always happens when their analogous relations appear.

\begin{table}[t]
\centering
\small
\scalebox{1}{
    \begin{tabular}{lcccccc}
    \toprule
    \multirow{2}{*}{\textbf{Models}} & \multicolumn{3}{c}{\textbf{FewRel}} & \multicolumn{3}{c}{\textbf{TACRED}} \\
    \cmidrule(r){2-4} \cmidrule(r){5-7}
    & G1 & G2 & G3 & G1 & G2 & G3 \\
    \midrule
    EMAR & 95.6 & 73.9 & 49.3 & 69.8 & 46.1 & 30.1 \\
    RP-CRE & 94.7 & 75.7 & 51.6 & 75.3 & 48.5 & 36.6 \\
    \midrule
    Sup. & 99.7 & 95.6 & 86.8 & 94.0 & 84.2 & 68.9 \\
    \bottomrule
    \end{tabular}
}
\caption{Retrieval precision of different methods on two benchmarks. We divide all relations into three groups according to their forgetting rate the same as Table \ref{tab:pilot_error}.}
\label{tab:retrival}
\end{table}

\begin{figure*}[t]
    \centering
    \scalebox{0.98}{
        \includegraphics[width=\linewidth]{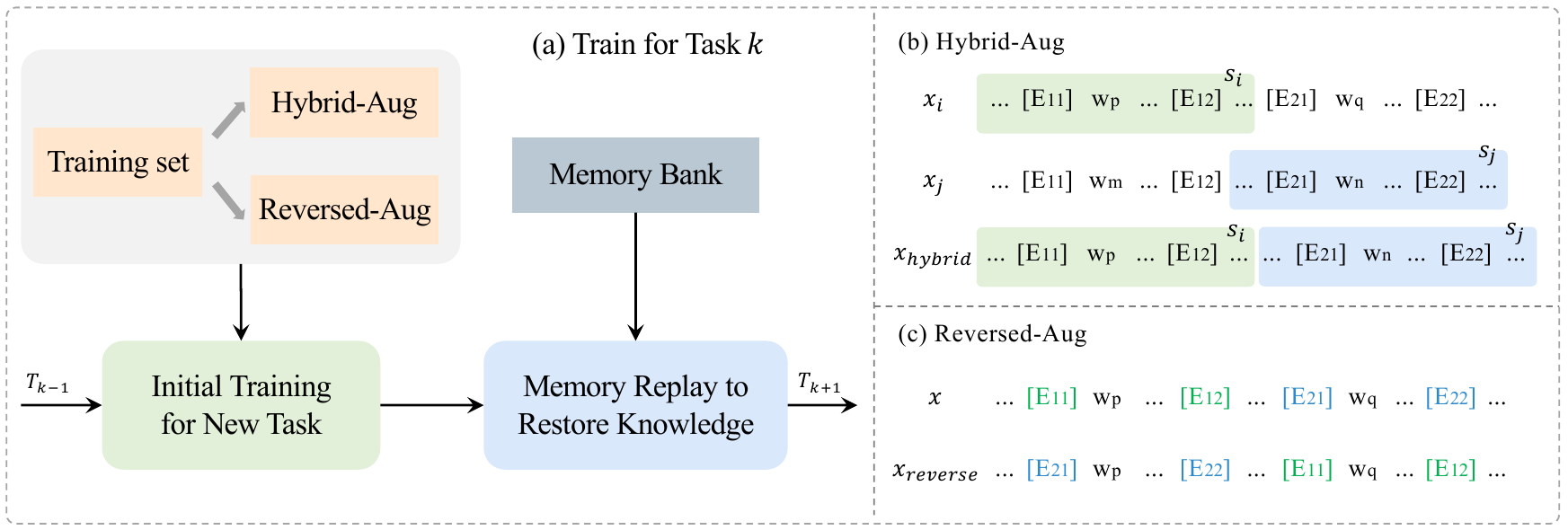}
    }    \caption{ (a) A demonstration for learning process of existing typical CRE models with our adversarial class augmentation mechanism. (b) Hybrid-class augmentation. (c) Reversed-class augmentation. We use ``[E$_{11}$]/[E$_{12}$]'' and ``[E$_{21}$]/[E$_{22}$]'' to mark the head entity $e^1$ and tail entity $e^2$, respectively.}
    \label{fig:method}
\end{figure*}

To confirm our hypothesis, we propose a retrieval test:
after the CRE model is trained to identify a specific relation $r$,
we use the trained model to retrieve instances of $r$ from the whole test set according to the similarity of  representations\footnote{Please refer to Appendix \ref{app:retrieval} for details of our retrieval test.}.
If the learned representations are not robust enough, the corresponding retrieval precision will be relatively low.
Table \ref{tab:retrival} shows the result of the retrieval results of two CRE models and the supervised model.
Compared with the supervised model, two CRE methods retrieve much more unrelated instances, especially for relations suffering severe forgetting, showing that the CRE models indeed learn representations that lack robustness.

\section{Methodology}

Recently, adversarial data augmentation has shown promise for avoiding models from learning spurious shortcuts in the easy dataset \cite{Volpi_2018_adv, zhao_2020_maximumentropy, hendrycks_2020_augmix,zhu_2021_dualaug}.
Therefore, in this section, 
we propose a simple yet effective adversarial class augmentation mechanism (\methodname{}) containing two kinds of class augmentation to help the CRE model learn more robust representations.
\begin{table*}[t]
    \centering
    % \small
    \scalebox{0.90}{
    \begin{tabular}{lcccccccccc}
     \toprule
     \multicolumn{11}{c}{\textbf{FewRel}} \\
     \midrule
      \textbf{Models} & \textbf{T1} & \textbf{T2} & \textbf{T3} & \textbf{T4} & \textbf{T5} & \textbf{T6} & \textbf{T7} & \textbf{T8} & \textbf{T9} & \textbf{T10} ($\Delta$)  \\
    \midrule
     EA-EMR \cite{wang2019eamar}  & 89.0 & 69.0 & 59.1 & 54.2 & 47.8 & 46.1 &  43.1 & 40.7 & 38.6 & 35.2 ( \ \ -- \ \ ) \\
     CML \cite{wu2021curriculum} & 91.2 & 74.8 & 68.2 & 58.2 & 53.7 & 50.4 & 47.8 & 44.4 & 43.1 & 39.7 ( \ \ -- \ \ ) \\
    RPCRE \cite{cui2021rpcre} & 97.9 & 92.7 & 91.6 & 89.2 & 88.4 & 86.8 & 85.1 & 84.1 & 82.2 & 81.5 ( \ \ -- \ \ )  \\
    CRL \cite{zhao2022crl} & 98.2 & 94.6 & 92.5 & 90.5 & 89.4 & 87.9 & 86.9 & 85.6 & 84.5 & 83.1 ( \ \ -- \ \ ) \\
    \midrule
    RP-CRE$^{\dagger}$ \cite{cui2021rpcre} & 97.8 & 94.7 & {92.1} & 90.3 & 89.4 & 88.0 & 87.1 & 85.8 & 84.4 & 82.8 ( \ \ -- \ \ ) \\
    RP-CRE$^{\dagger}$ + \textbf{\methodname{}} & 98.0 & 94.7 & 92.2 & 90.6 & 89.9 & 89.0 & 87.5 & 86.4 & 85.7 & 83.8 ($\uparrow$ 1.0)\\
    \midrule
    EMAR$^{\dagger}$ \cite{han2020emar} & 98.1 & 94.3  & 92.3 & 90.5 & 89.7 & 88.5 & 87.2 & 86.1 & 84.8 & 83.6 ( \ \ -- \ \ ) \\
    EMAR$^{\dagger}$ + \textbf{\methodname{}} & \bf{98.3} & \bf{95.0}  & \bf{92.6} & \bf{91.3} & \bf{90.4} & \bf{89.2} & \bf{87.6} & \bf{87.0} & \bf{86.3} & \textbf{84.7} ($\uparrow$ 1.1) \\
     \bottomrule
     \toprule
    \multicolumn{11}{c}{\textbf{TACRED}} \\
    \midrule
      \textbf{Models} & \textbf{T1} & \textbf{T2} & \textbf{T3} & \textbf{T4} & \textbf{T5} & \textbf{T6} & \textbf{T7} & \textbf{T8} & \textbf{T9} & \textbf{T10} ($\Delta$) \\
    \midrule
    EA-EMR \cite{wang2019eamar} & 47.5 & 40.1 & 38.3 & 29.9 & 24.0 & 27.3 &  26.9 & 25.8 & 22.9 & 19.8 ( \ \ -- \ \ )\\
        CML \cite{wu2021curriculum} & 57.2 & 51.4 & 41.3 & 39.3 & 35.9 & 28.9 & 27.3 & 26.9 & 24.8 & 23.4 ( \ \ -- \ \ )\\
      RP-CRE \cite{cui2021rpcre} & 97.6 & 90.6 & 86.1 & 82.4 & 79.8 & 77.2 & 75.1 & 73.7 & 72.4 & 72.4 ( \ \ -- \ \ )\\
        CRL \cite{zhao2022crl} & 97.7 & 93.2 & 89.8 & 84.7 & 84.1 & 81.3 & \bf{80.2} & \bf{79.1} & \bf{79.0} & 78.0 ( \ \ -- \ \ )\\
      \midrule
      RP-CRE$^{\dagger}$ \cite{cui2021rpcre} & 97.5 & 92.2 & 89.1 & 84.2 & 81.7 & 81.0 & 78.1 & 76.1 & 75.0 & 75.3 ( \ \ -- \ \ )\\
      RP-CRE$^{\dagger}$ + \textbf{\methodname{}} & 97.8 & \bf{93.6} & 89.9  & 84.4 & 82.7 & 81.1 & 78.2 & 77.7 & 75.5 & 76.2 ($\uparrow$ 0.9)\\
      \midrule
      EMAR$^{\dagger}$ \cite{han2020emar} & \bf{98.3} & 92.0 & 87.4 & 84.1 & 82.1 & 80.6 & 78.3 & 76.6 & 76.8 & 76.1 ( \ \ -- \ \ )\\
      EMAR$^{\dagger}$ + \textbf{\methodname{}} & 98.0 & 92.1 & \bf{90.6} & \bf{85.5} & \bf{84.4} & \bf{82.2} & 80.0 & 78.6 & 78.8 & \textbf{78.1} ($\uparrow$ 2.0)\\
    \bottomrule
    \end{tabular}}
    \caption{Accuracy (\%) on all seen relations at the stage of learning current tasks. We report the average accuracy of $5$ different runs. $^{\dagger}$ denotes our reproduced results with the open codebases. Other results are directly taking from \citet{zhao2022crl}. $\Delta$ is the performance gap in T10 between original CRE models and the models with our adversarial class augmentation mechanism ( + \textbf{\methodname{}}). EMAR and RP-CRE with \methodname{} significantly outperform their corresponding vanilla models ($p < 0.05$).
    }
    \label{tab:main}
\end{table*}

\subsection{Two-Stage Training}
Our \methodname{} is model-agnostic and utilizes popular state-of-the-art CRE models as the backbone. Therefore, we first briefly introduce the two-stage training process of these CRE models.

CRE model aim to finish a sequence of tasks $(T_1, T_2, ..., T_k)$.
Without loss of generality, 
% in the $i$-th task $T_i$, 
we represent CRE model with two components:
\textbf{1)} an encoder, which maps an input instance $x$ into a representation vector;
\textbf{2)} a classifier, which produces a probability distribution over all seen relations till current task as the prediction for $x$.
% \textbf{1)} an encoder $f_{\Theta}: \mathcal{X} \to \mathcal{Z}$, which maps an input text $x$ into a representation vector $z=f_{\Theta}(x)\in\mathbb{R}^d$;
% \textbf{2)} a classifier $g_{\theta_t}: \mathcal{Z} \to \mathbb{R}^{|C_i|}$, which produces a probability distribution as the prediction for $x$, where $|C_i|$ is the number of seen relations till $T_i$. 
As shown in Figure \ref{fig:method}(a),
previous CRE methods \cite{han2020emar,cui2021rpcre,zhao2022crl,wang2022less} can be generally formulated as a two-stage training process.
\textbf{1)} initial training:
they first expand the class node in the classifier for new relations, and then train the CRE model with only new data to learn new relations;
\textbf{2)} memory replay:
they first update the memory bank with new data, and then replay the memory bank to restore the knowledge of previously learned relations.
Specifically, previous work
% nearly do not pay attention to the initial training stage, and just simply train the model with new data. 
mainly focuses on the memory replay stage and proposes various sophisticated mechanisms 
% such as relation prototypes \cite{han2020emar}, memory refinement network \cite{cui2021rpcre}, contrastive replay and knowledge distillation \cite{zhao2022crl} 
to better retain or recover the learned knowledge, while improvements to the initial training stage remain under-explored.
See more details of these CRE methods in their original paper.

\subsection{Adversarial Class Augmentation}
Orthogonal to all previous CRE models, our ACA instead focuses on the first initial training stage to improve the robustness of newly learned relation representations.
% , which can better against the appearance of analogous relations in the future.
Specifically, when a new task $T_i$ comes, ACA first augments the new relations $R_i$ based on the new training set $D_i$, and then trains the original relations and synthesized classes together.

\paragraph{Hybrid-class Augmentation}
Given $N$ new relations, we pair them randomly and get $\lfloor N / 2 \rfloor$ relation pairs.
We construct hybrid synthetic classes based on these relation pairs.
Specifically, for a relation pair $\{r_i, r_j\}$ with the relations $r_i$ and $r_j$, we use two instances, $x_i$ from $r_i$ and $x_j$ from $r_j$, to generate a hybrid instance $x_{hybrid}$ for the extra synthetic class $r_{ij}$.
As shown in Figure \ref{fig:method}(b), we first extract a span $s_i$ that contains the head entity $e_i^1$ but excludes the tail entity $e_i^2$ from $x_i$, and a span $s_j$ that contains the tail entity $e_j^2$ but excludes the head entity $e_j^1$ from $x_j$, and then concatenate $s_i$ and $s_j$ to form $x_{hybrid}=[s_i;s_j]$.
Through the hybrid-class augmentation, we can construct $\lfloor N / 2 \rfloor$ extra classes for the current new task.

\paragraph{Reversed-class Augmentation}
We classify all relations into two categories, symmetric and asymmetric relations.
The symmetric relation means that the order of the head and tail entities does not matter, e.g, ``sibling'' and ``spouse'' (please refer to Appendix \ref{app:rev} for details of symmetric relations on two datasets).
In contrast, the semantic of the asymmetric relations is related to the choice of head and tail entities, e.g., ``located in'' and ``mother''.
As shown in Figure \ref{fig:method}(c), reversed-class augmentation reverses the head and tail entities of each asymmetrical relation to construct the extra classes.

\paragraph{Adversarial Training} 
Given $N$ new relations, we can generate at most $N+\lfloor N / 2 \rfloor$ hard negative classes using the two augmentation methods.
Thus, in the initial training stage, the model is jointly trained to classify $(2N+\lfloor N / 2 \rfloor)$ classes to better learn the original new relations.
At the end of initial training, the extended class nodes of augmented classes in the classifier will be removed.

\section{Experiments}

\subsection{Experimental Setups}
\label{exp_setup}
\paragraph{Datasets}
Following previous works \cite{han2020emar,wu2021curriculum,cui2021rpcre, zhao2022crl}, our experiments are conducted upon two widely used datasets, \textbf{FewRel} \cite{han2018fewrel} and \textbf{TACRED} \cite{zhang2017tacred}. Please refer to Appendix \ref{app:data} for details of these two datasets.
We construct $5$ different task sequences for both FewRel and TACRED.
For each task sequence, we simulate 10 tasks by randomly dividing all relations of the dataset into 10 sets.
For a fair comparison, our $5$ task sequences are exactly the same as that of \citet{cui2021rpcre} and \citet{zhao2022crl}.

\paragraph{Evaluation Metrics}
Following \citet{cui2021rpcre} and  \citet{zhao2022crl}, we use average accuracy on all seen tasks as our evaluation metric. 
% because average accuracy can better measure the effect of catastrophic forgetting.
For a stronger method, the average accuracy of each task should be consistently higher than that of baselines.

\paragraph{Baselines}
We consider the following baselines:
\textbf{EA-EMR} \cite{wang2019eamar}, which maintains a memory replay and embedding alignment mechanism to alleviate catastrophic forgetting;
\textbf{CML} \cite{wu2021curriculum}, which introduces curriculum learning and meta-learn  to alleviate catastrophic forgetting in CRE;
\textbf{EMAR} \cite{han2020emar}, which proposes a memory activation and reconsolidation mechanism to retain the learned knowledge.
Note that the original EMAR was based on a Bi-LSTM encoder, and we re-implement EMAR with BERT;
\textbf{RP-CRE} \cite{cui2021rpcre}, which proposes a memory network to retain the learned representations with relation prototypes;
\textbf{CRL} \cite{zhao2022crl}, which adopts contrastive learning replay and knowledge distillation to retain the learned knowledge.

\paragraph{Implement Details} Our ACA is model-agnostic, and we choose two state-of-the-art CRE models, EMAR and RP-CRE as our backbone to evaluate \methodname{}.
The number of stored instances of each relation in the memory bank is $10$.
All hyperparameters of EMAR and RP-CRE are the same as that of their origin paper.
\methodname{} does not introduce any model hyperparameters.
We run our code on a single NVIDIA A40 GPU with 48GB memory, and report the average result of $5$ different task sequences.

\subsection{Main Results}
The performances of our \methodname{} and baselines are shown in Table \ref{tab:main}.
As shown, after applying \methodname{},
the performances of EMAR and RP-CRE consistently improve in nearly all training stages of two benchmarks.
Previous CRE work usually regards the accuracy of the last task as the most important metric.
For the accuracy of T10, 
our proposed \methodname{} improves RP-CRE/EMAR by $1.0/1.1$ and $0.9/2.0$ accuracy on FewRel and TACRED, respectively.
Furthermore, EMAR+\methodname{} achieves new state-of-the-art results on both two benchmarks.
These results demonstrate the effectiveness and universality of our proposed method.

\section{Analysis}
\begin{table}[t]
\centering
\small
\scalebox{1}{
    \begin{tabular}{lcc}
    \toprule
    \textbf{Models} & \textbf{FewRel} & \textbf{TACRED}   \\
    \midrule
    EMAR+\methodname   & 84.7 & 78.1   \\
    \midrule
    w/o hybrid-class aug. & 84.3 &  77.4 \\
    w/o reversed-class aug. & 83.9 & 76.8\\
    w/o both classes aug. & 83.6 & 76.1 \\
    \bottomrule
    \end{tabular}
}
\caption{The effect of two class augmentation methods on two benchmarks. ``aug.'' is short for augmentation.}
\label{tab:ablation}
\end{table}

\subsection{Ablation Study}
To further explore the effectiveness of our proposed two class augmentation methods, we conduct an ablation study.
% by removing one augmentation at a time.
Table \ref{tab:ablation} shows the results of EMAR with different augmentation methods on two benchmarks.
We find that both augmentations are conducive to the model performance, and they are complementary to each other.
In addition, the reversed-class augmentation is more effective than the hybrid-class augmentation.
We think the reason is that the reversed-class augmentation can maintain the fluency of the constructed sentences, while the hybrid-class augmentation cannot.

\subsection{Robust Representation Learning}
\begin{table}[t]
\centering
\small
\scalebox{0.9}{
    \begin{tabular}{lcccccc}
    \toprule
    \multirow{2}{*}{\textbf{Models}} & \multicolumn{3}{c}{\textbf{FewRel}} & \multicolumn{3}{c}{\textbf{TACRED}} \\
    \cmidrule(r){2-4} \cmidrule(r){5-7}
    & G1 & G2 & G3 & G1 & G2 & G3 \\
    \midrule
    EMAR & 95.6 & 73.9 & 49.3 & 69.8 & 46.1 & 30.1  \\
    EMAR+\methodname & 97.3 & 83.0 & 57.7 & 77.9 & 52.6 & 33.5 \\
    \bottomrule
    \end{tabular}
}
\caption{Retrieval precision of different models on FewRel. The results are divided into three groups according to their forgetting rate of EMAR.}
\label{tab:ret-ahead}
\end{table}

\begin{figure}[t]
    \centering
    \includegraphics[width=\linewidth]{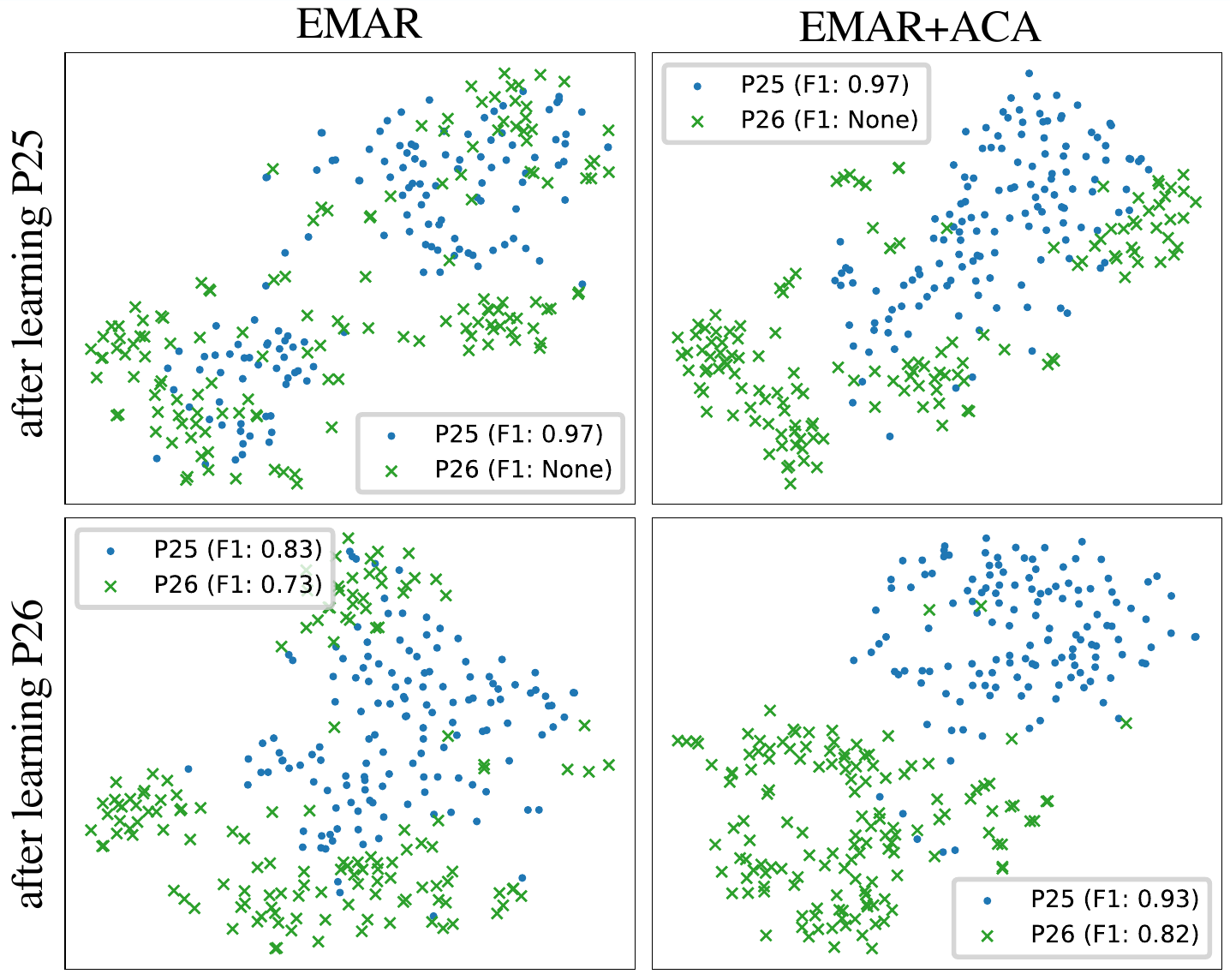}
    \caption{The representation of instances belonging to P25 (``mother'') and P26 (``spouse'') after learning P25 and P26, respectively.}
    \label{fig:rep}
\end{figure}

\begin{figure*}[t]
    \centering
    \subfloat[Results on FewRel]{
    % \label{fig:layers:a}
        \includegraphics[width=0.49\linewidth]{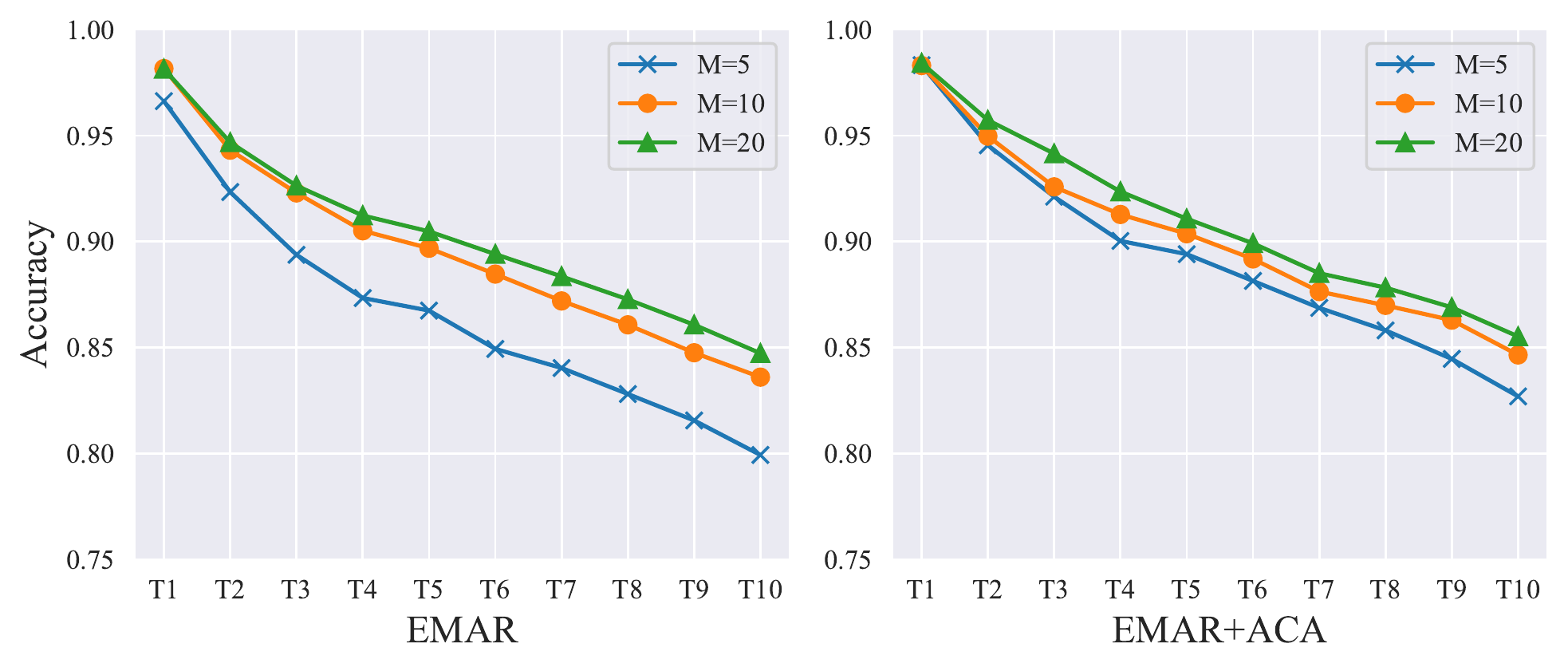}
       % \caption{}
    }
    \subfloat[Results on TACRED]{
    % \label{fig:layers:b}
        \includegraphics[width=0.49\linewidth]{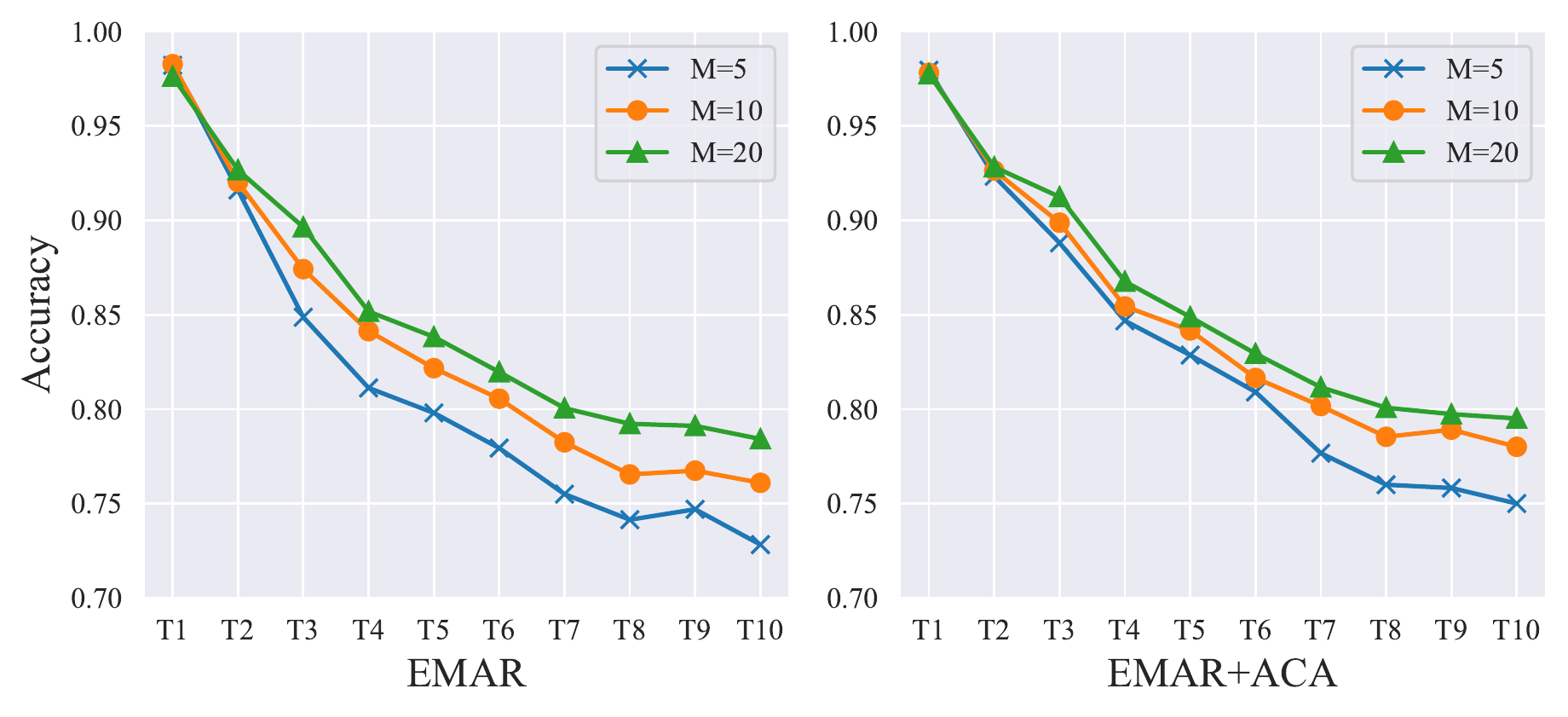}
        % \caption{}
    }
    \caption{
    Comparison of model's dependence on memory size. We report the average results of $5$ different runs. ACA improves the stability of EMAR on two benchmarks.
    }
    \label{fig:memory}
\end{figure*}

Our proposed \methodname{} aims to learn robust representations that can better distinguish analogous relations.
To further confirm the effectiveness of our method, we first reproduce the retrieval test introduced in our pilot experiments (see Appendix \ref{app:retrieval} for more details).
Table \ref{tab:ret-ahead} shows the results of EMAR and EMAR+\methodname{} on two benchmarks.
As is shown, \methodname{} can significantly increase the precision of retrieval results, showing that our method indeed helps the model learn more robust representations.

We also conduct a case study to intuitively show the effectiveness of our method.
We consider two analogous relations, P25 (``mother'') and P26 (``spouse'')\footnote{Please refer to Appendix \ref{app:rp} for more cases.}, and EMAR catastrophically forgets P25 when P26 appears.
We use t-SNE to visualize the representation of all instances belonging to these two relations after the model learning P25 and P26, respectively.
As shown in Figure \ref{fig:rep}:
\textbf{1)} For EMAR, after the model learns the relation P25, it relies on spurious cues, such as entity types, to identify instances of P25.
% , and thus the learned representations can also represent instances of P26.
% which can also represent P26.
Therefore, the representations of the instances belonging to P25 and P26 are mixed together, which means the learned representation of P25 can also represent instances of P26.
When P26 appears, it is hard to learn a more robust representation of P25 with P26 with only very limited memory instances of P25, and thus EMAR catastrophically forgets P25 (the F1 score of P25 significantly degrades $14$ points);
\textbf{2)} For EMAR+\methodname{}, when learning P25, the model can learn more robust representations of P25 with our augmented relations. 
Thus, the representation of instances belonging to P25 and P26 is much more separable than that of EMAR. 
When P26 appears, the forgetting problem of P25 is greatly alleviated (the F1 score of P25 only drops $4$ points).

\subsection{Influence of Memory Size}

Memory size is the number of memorized instances for each relation, which is a key factor for the model performance of rehearsal-based CRE methods.
Therefore, in this section, we study the influence of memory size on our ACA.

We compare the performance of EMAR and EMAR+\methodname{} with memory sizes $5$, $10$ and $20$.
As shown in Figure \ref{fig:memory}:
\textbf{1)} As the size of the memory decreases, the performances of both models drop, showing the importance of the memory size for CRE models;
\textbf{2)} On both FewRel and TACRED, EMAR+\methodname{} outperforms EMAR under all three different memory sizes, further demonstrating the effectiveness of our ACA;
\textbf{3)} As memory size decreases, EMAR+\methodname{} shows a relatively stable performances.
Specifically, EMAR+\methodname{} outperforms EMAR $2.8$,  $1.1$ and $0.7$ accuracy on FewRel when the memory size is $5$, $10$, $20$, respectively. A similar tendency is also observed on TACRED.
These results further demonstrate the effectiveness of robust representations learned through our ACA.

\subsection{Error Analysis}
\begin{table}[t]
\centering
\small
\scalebox{1}{
    \begin{tabular}{ccccccc}
    \toprule
    \multicolumn{3}{c}{\textbf{\textsc{Group}}}  
    & \multicolumn{2}{c}{\textbf{EMAR}} & \multicolumn{2}{c}{\textbf{+\methodname}}  \\
  \cmidrule(r){1-3}   \cmidrule(r){4-5}  \cmidrule(r){6-7}
    \textbf{ID} & \textbf{MS} & \textbf{F1$^*$} & \textbf{FR} &\textbf{F1}  & \textbf{FR} &\textbf{F1} ($\Delta$) \\
    \midrule
    G1  &  0.39 & 95.0  & 2.5 & 91.2 & 2.6 & 91.4 ($\uparrow 0.2$)\\
    G2  &  0.51 & 90.8  & 5.2 & 83.3 & 4.9 & 84.8 ($\uparrow 0.5$)\\
    G3  &  0.67 & 84.0  & 7.5 & 75.3 & 7.1 & 76.2 ($\uparrow 0.9$)\\
    \bottomrule
    \end{tabular}
}
\caption{We equally divide relations of FewRel into $3$ groups by their \textit{max similarity} (MS). F1$^*$ is the F1 score of the supervised model. $\Delta$ denotes the performance gap between EMAR and EMAR+ACA.}
\label{tab:error}

\end{table}

In this section, we conduct an error analysis to show the effectiveness of \methodname{} and the challenge of CRE.
Through our analysis of catastrophic forgetting, we find that the performance of relations is highly related to their \textit{max similarity}.
Therefore, we equally divide the relations into three groups according to their \textit{max similarity}.
As shown in Table \ref{tab:error}: 
\textbf{1)} \methodname{} mainly improves the performance and reduces the forgetting rate of relations with large \textit{max similarity};
\textbf{2)} although \methodname{} is efficient, relations with large \textit{max similarity} still suffer from catastrophic forgetting and have a large performance gap with the supervised model.
Therefore, future work should pay more attention to these relations.

\section{Conclusion}
In this paper, we conduct a series of empirical study to analyze catastrophic forgetting in CRE, and observe that catastrophic forgetting mostly happens on some specific relations, and significant performance degradation tends to occur when their analogous relations appear in subsequent tasks.
Based on our observations, we find an important reason for catastrophic forgetting in CRE that all previous works overlooked, i.e., the CRE models suffer from learning shortcuts to identify new relations, which are not robust enough against the appearance of their analogous relations.
To this end, we propose a simple yet effective adversarial class augmentation mechanism to help CRE models learn more robust representations.
Extensive experiments on two benchmarks show that our method can further improve the performance of two state-of-the-art CRE models.

\section*{Limitations}
Our paper has several limitations:
\textbf{1)} Although we provide a new perspective from the shortcut learning to explain catastrophic forgetting, and utilize a retrieval test to confirm our hypothesis, we do not explore which types of shortcuts are learned by CRE models;
\textbf{2)} Our ACA with two class augmentation methods is specially designed for CRE.
However, our findings about catastrophic forgetting in this paper may be common in the context of continual learning.
Therefore, it would be better if we can propose more universal adversarial training methods which can be adapted to all continual learning systems;
\textbf{3)} ACA conducts the class augmentation before the initial training stage, which introduces extra computational overhead on top of backbone CRE models.

\section*{Acknowledgements}
We thank Dawei Zhu and Zhejian Zhou for proofreading the paper and providing insightful comments. 
We thank Lei Li for some useful discussions.
We also thank the anonymous reviewers for their thoughtful and constructive comments.
This work is supported by National Key Research and Development Project (2019YFB1704002) and NSFC project U19A2065.

\bibliography{anthology,custom}
\bibliographystyle{acl_natbib}

\appendix

\clearpage

\section{Forgetting Rate on TACRED}
\label{app:pilot_tacred}

We show the performance of two strong baselines on TACRED in Table \ref{tab:pilot_tacred}.
We also divide relations in TACRED into three groups according to their forgetting rate.

\section{Cases for performance curves of different relations}
\label{app:sud_drop}

\begin{figure*}[t]
    \centering
    \scalebox{0.9}{
        \includegraphics[width=\linewidth]{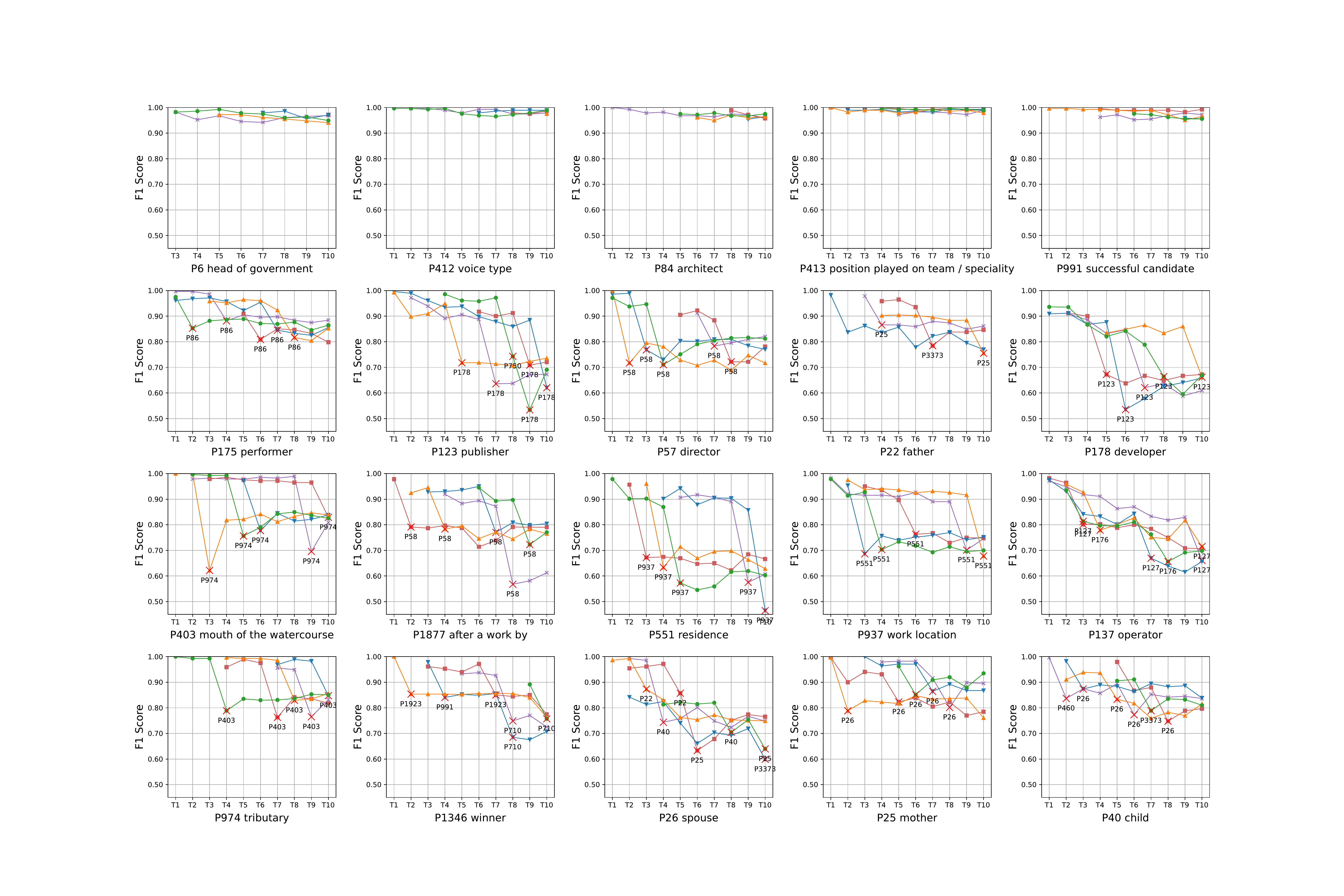}
    }
    \caption{Cases of performance curves of different relations in FewRel. For each relation, we illustrate its F1 curves corresponding to five different task sequences. Note that the performance of some relations hardly degrades, while other relations suffer from catastrophic forgetting. For a specific relation $r$, we also plot bad cases (the F1 curve has a degradation more than $0.1$ F1 score) containing top-$5$ most similar relations of $r$.}
    \label{app:cf_case}
\end{figure*}

As illustrated in Figure \ref{app:cf_case}, we provide cases to illustrate catastrophic forgetting only appears on some specific relations and significant performance degradation always occurs when analogous relation appears.
For each relation, we plot the performance curves corresponding to five different task sequences.
We notice that some relations almost have no performance degradation during the training process (as shown in the top row of Figure \ref{app:cf_case}), while some relations suffer from catastrophic forgetting (as shown in the bottom three rows of Figure \ref{app:cf_case}).
We further observe that when the performance curve of a specific relation $r$ has a sudden degradation, the corresponding task always contains relations very similar to $r$.

\section{Retrieval Test}
\label{app:retrieval}
As discussed in Section \ref{sec:reason}, a potential reason for catastrophic forgetting in CRE is the model only learns the spurious shortcuts in the continual learning setting.
In order to evaluate the representation ability of the CRE model, we propose a retrieval test analysis.

Given an instance $x$, a CRE method utilizes an encoder $f$ to encode its semantic features for learning and classifying relations,
\begin{equation}
    \bm{h}=f(x).
\end{equation}
For a relation, if its F1 score degrades greater than 0.1 after the model learning a new task, we consider it as an relation suffering severe forgetting.
We group all relations suffering severe forgetting into a set $R_f$.
For a relation $r\in R_f$, we additionally randomly sample 7 relations $\{r^*_1, ..., r^*_7\}$ (3 relations for TACRED) from $R\setminus R_f$ to build a pseudo task $T^*$ containing instances from $R^*=\{r, r^*_1, ..., r^*_7\}$, where $R$ is the relation set of the entire dataset.
After training the CRE model on our built pseudo task $T^*$, we obtain the prototype $\bm{p}_r$ of the relation $r$, that is, the mean embedding of all instances belonging to $r$,
\begin{equation}
    \bm{p}_r=\frac{\sum^{|r|}_{j=1} f(x_j^r)}{|r|},
\end{equation}
where $|r|$ is the number of instances of relation $r$.
We also obtain embeddings of each instance in the entire test set,
\begin{equation}
    I=\{\bm{h}_i|\bm{h}_i=f(x_i),\ x_i\in \bigcup_{r_j \in R} Q_j\}.
\end{equation}
Then we compute the cosine similarity between $\bm{p}_r$ and $\bm{h}_i\in I$:
\begin{equation}
    \mathop{\rm{Sim}}(\bm{p}_r, \bm{h}_i)=\frac{\bm{p}_r \cdot \bm{h}_i}{|\bm{p}_r|\cdot|\bm{h}_i|},
\end{equation}

We consider rank-based metrics and use the mean precision at k (\textit{P@k}), which is the proportion of instances whose label is $r$ in the top-k similar set.
Specifically, for FewRel, we use \textit{P@100} as metric.
For TACRED, because this dataset has a severe imbalance problem and some relations only have less than 50 instances, we use mean \textit{P@$\left| Q_r\right|$} as metric, where $\left| Q_r\right|$ is the size of test set corresponding to relation $r$.
If the retrieval precision is high, we can say that the model learns robust representations.

\begin{table}[t]
\centering
\small
\scalebox{0.95}{
    \begin{tabular}{lccccl}
    \toprule
    \textbf{Model} & \textbf{ID}   & \textbf{FR (\%)} & \textbf{MS} & \textbf{F1} & \textbf{F1$^*$} ($\Delta$) \\
    \midrule
    \multirow{3}{*}{\rotatebox[origin=c]{45}{EMAR}} 
   & G1  & 2.2 & 0.49 & 93.5 & 95.7 ({$\uparrow$ 2.2}) \\
    & G2  &  7.1 & 0.64 & 76.5 & 85.8 ({$\uparrow$ 9.3})  \\
    & G3  &  13.7 & 0.75 & 56.9 & 68.0 ({$\uparrow$ 11.1})  \\
    \midrule
   \multirow{3}{*}{\rotatebox[origin=c]{45}{RP-CRE}} 
   & G1  & 2.3 & 0.49 & 93.2 & 95.7 ({$\uparrow$ 2.5}) \\
    & G2  &  7.5 & 0.62 & 76.8 & 86.0 ({$\uparrow$ 9.2})  \\
    & G3  &  15.3 & 0.76 & 56.2 & 67.8 ({$\uparrow$ 11.6})  \\
    \bottomrule
    \end{tabular}
}
\caption{ We divide relations of TACRED into $3$ groups according to their forgetting rate (FR). ``MS'' is the short for \textit{max similarity}. F1 and F1$^*$ are the Macro-F1 scores of EMAR/RP-CRE and the supervised model, respectively. $\Delta$ is the performance gap between $2$ CRE models and the supervised model.}
\label{tab:pilot_tacred}
\end{table}

\section{Datasets}
\label{app:data}
Following previous work \cite{han2020emar,wu2021curriculum,cui2021rpcre, zhao2022crl}, our experiments are conducted upon the following two widely datasets, and the training-test-validation split ratio is 3:1:1:
\paragraph{FewRel} \cite{han2018fewrel} It is a relation extraction dataset originally proposed for few-shot learning, which contains 80 relations, each with 700 instances.
Following \citet{cui2021rpcre,zhao2022crl}, we use the training and validation set of FewRel for experimental.
\paragraph{TACRED} \cite{zhang2017tacred} It is a large-scale RE dataset built on news networks and online documents, containing 42 relations (including \textit{no\_relation}) and 106264 samples.
Following \citet{cui2021rpcre}, \textit{no\_relation} was removed in our experiments, and the number of training samples for each relation is limited to 320 and the number of test samples of each relation to 40.

\section{Symmetric Relations in Two Datasets}
\label{app:rev}
In our reversed-class augmentation, we divide all relations into two categories, symmetric relation and asymmetric relation.
The symmetric relation denotes the relation semantic is independent of which of the two given entities is the head or tail entity, and the relations except symmetric relations are asymmetric relations.
(1) In FewRel, we find $2$ symmetric relations, ``P26 (spouse)'' and ``P3373 (sibling)''.
(2) In TACRED, we find $5$ symmetric relations, ``per:siblings'', ``org:alternate names'', ``per:spouse'', ``per:alternate names'' and ``per:other family''.

\section{Robust Representation}
\label{app:rp}
\begin{figure*}[t]
    \centering
    \scalebox{0.9}{
        \includegraphics[width=\linewidth]{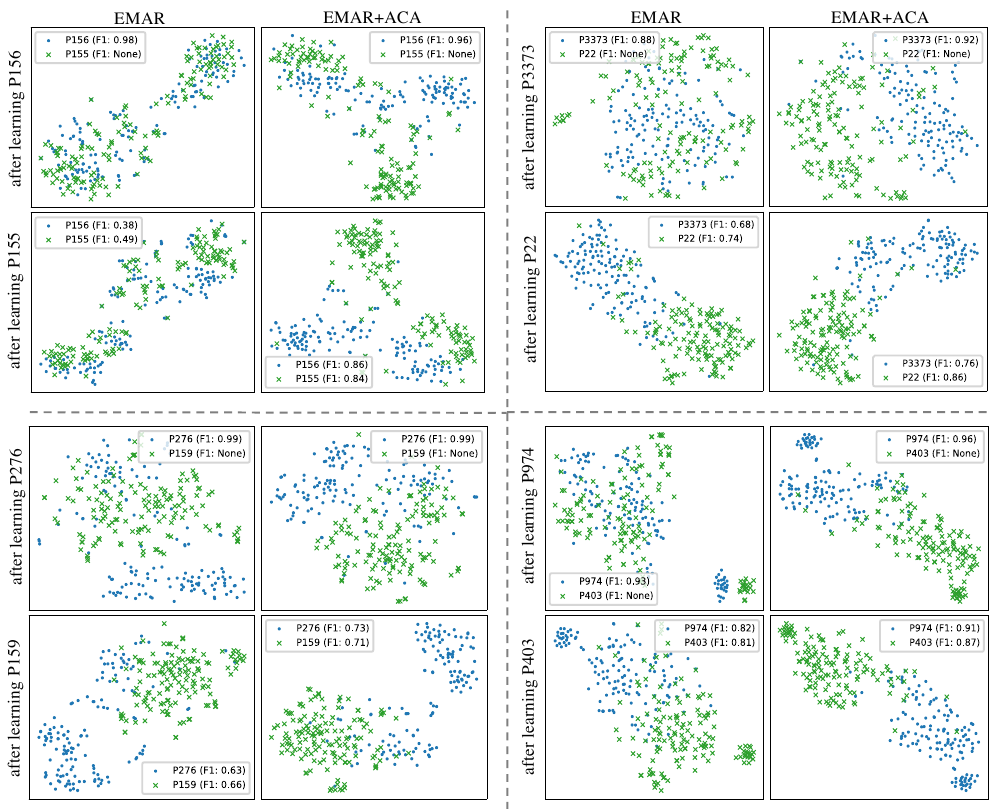}
    }
    \caption{More cases of robust representation learning.}
    \label{app:rep}
\end{figure*}

In this section, we provide more cases to intuitively show the effectiveness of our look-ahead learning for learning robust representation. Please refer to Figure \ref{app:rep} for details.

\section{Relations in Different Groups}
\label{rdg}

As discussed in Section \ref{sec:pilot}, we divide all relations into three equal-sized groups based on their FR from small to large.
In this section, we show example relations in Group 1 and Group 3 of FewRel in Table \ref{tab:similarity_case}.
It is easy to see that relations in Group 3, which suffer from significant performance degradation, have larger similarity to other relations in the dataset than relations in Group 1.

\begin{table*}[t]
\centering
\small
\scalebox{0.85}{
    \begin{tabular}{l|l|l}
    \toprule
    \multicolumn{3}{c}{\textbf{Example Relations of Group 1}} \\
    \midrule
    \textbf{Wikidata ID} & \textbf{Relation Name} & \textbf{Top-3 Most Similar Relations}   \\
    \midrule
    \multirow{3}*{P59} & \multirow{3}*{constellation} & (0.693) P106, occupation\\
    ~ & ~ & (0.673) P27, country of citizenship\\
    ~ & ~ & (0.663) P641, sport\\
    \midrule
    \multirow{3}*{P1411} & \multirow{3}*{nominated for} & (0.758) P449, original network\\
    ~ & ~ & (0.757) P750, distributor\\
    ~ & ~ & (0.755) P27, country of citizenship\\
    \midrule
    \multirow{3}*{P2094} & \multirow{3}*{competition class} & (0.789) P641, sport\\
    ~ & ~ & (0.780) P463, member of\\
    ~ & ~ & (0.755) P27, country of citizenship\\
    \midrule
    \multirow{3}*{P105} & \multirow{3}*{taxon rank} & (0.806) P31, instance of\\
    ~ & ~ & (0.794) P361, part of\\
    ~ & ~ & (0.784) P206, located in or next to body of water\\
    \midrule
    \multirow{3}*{P1435} & \multirow{3}*{heritage designation} & (0.846) P177, crosses\\
    ~ & ~ & (0.823) P31, instance of\\
    ~ & ~ & (0.822) P131, located in the administrative territorial entity\\
    \midrule
    \multirow{3}*{P1344} & \multirow{3}*{participant of} & (0.878) P27, country of citizenship\\
    ~ & ~ & (0.857) P463, member of\\
    ~ & ~ & (0.850) P551, residence\\
    \midrule
    \multirow{3}*{P410} & \multirow{3}*{military rank} & (0.886) P39, position held\\
    ~ & ~ & (0.882) P241, military branch\\
    ~ & ~ & (0.848) P22, father\\
    \midrule
    \multirow{3}*{P84} & \multirow{3}*{architect} & (0.890) P6, head of government\\
    ~ & ~ & (0.889) P127, owned by\\
    ~ & ~ & (0.875) P86, composer\\
    \midrule
    \multirow{3}*{P306} & \multirow{3}*{operating system} & (0.891) P400, platform\\
    ~ & ~ & (0.881) P178, developer\\
    ~ & ~ & (0.856) P31, instance of\\
    \midrule
    \multirow{3}*{P1303} & \multirow{3}*{instrument} & (0.894) P101, field of work\\
    ~ & ~ & (0.893) P463, member of\\
    ~ & ~ & (0.884) P106, occupation\\
    \bottomrule
    \toprule
    \multicolumn{3}{c}{\textbf{Example Relations of Group 3}} \\
    \midrule
    \textbf{Wikidata ID} & \textbf{Relation Name} & \textbf{Top-3 Most Similar Relations}   \\
    \midrule
    \multirow{3}*{P155} & \multirow{3}*{follows} & (0.991) P156, followed by\\
    ~ & ~ & (0.962) P361, part of\\
    ~ & ~ & (0.959) P527, has part\\
    \midrule
    \multirow{3}*{P706} & \multirow{3}*{located on terrain feature} & (0.984) P206, located in or next to body of water\\
    ~ & ~ & (0.966) P131, located in the administrative territorial entity\\
    ~ & ~ & (0.961) P4552, mountain range\\
    \midrule
    \multirow{3}*{P57} & \multirow{3}*{director} & (0.984) P58, screenwriter\\
    ~ & ~ & (0.974) P1877, after a work by\\
    ~ & ~ & (0.954) P86, composer\\
    \midrule
    \multirow{3}*{P22} & \multirow{3}*{father} & (0.981) P40, child\\
    ~ & ~ & (0.974) P26, spouse\\
    ~ & ~ & (0.972) P3373, sibling\\
    \midrule
    \multirow{3}*{P123} & \multirow{3}*{publisher} & (0.978) P178, developer\\
    ~ & ~ & (0.953) P750, distributor\\
    ~ & ~ & (0.943) P127, owned by\\
    \midrule
    \multirow{3}*{P127} & \multirow{3}*{owned by} & (0.977) P355, subsidiary\\
    ~ & ~ & (0.967) P137, operator\\
    ~ & ~ & (0.958) P159, headquarters location\\
    \midrule
    \multirow{3}*{P25} & \multirow{3}*{mother} & (0.976) P26, spouse\\
    ~ & ~ & (0.967) P40, child \\
    ~ & ~ & (0.961) P3373, sibling\\
    \midrule
    \multirow{3}*{P1877} & \multirow{3}*{after a work by} & (0.974) P58, screenwriter\\
    ~ & ~ & (0.958) P57, director\\
    ~ & ~ & (0.931) P86, composer\\
    \midrule
    \multirow{3}*{P17} & \multirow{3}*{country} & (0.968) P131, located in the administrative territorial entity\\
    ~ & ~ & (0.960) P361, part of\\
    ~ & ~ & (0.950) P159, headquarters location\\
    \midrule
    \multirow{3}*{P551} & \multirow{3}*{residence} & (0.968) P937, work location\\
    ~ & ~ & (0.939) P27, country of citizenship\\
    ~ & ~ & (0.934) P159, headquarters location\\
    \bottomrule
    \end{tabular}
}
\caption{Example relations in Group 1 and Group 3 of FewRel. For each relation, we show its top-3 most similar relations with their corresponding cosine similarity.}
\label{tab:similarity_case}
\end{table*}

\end{document}